%% file: main.tex
\documentclass[runningheads]{llncs}

\usepackage{eccv}

\usepackage{eccvabbrv}

\usepackage{graphicx}
\usepackage{booktabs}

\usepackage[accsupp]{axessibility}  
\usepackage{ulem}
\usepackage{graphicx} 
\usepackage{float} 
\usepackage{algorithm}
\usepackage{algorithmic}
\usepackage{breqn}
\usepackage{threeparttable}
\usepackage{multirow}
\usepackage{wrapfig}
\usepackage{multirow}
\usepackage{caption}
\usepackage{ulem} 

\usepackage[pagebackref,breaklinks,colorlinks,citecolor=eccvblue]{hyperref}

\usepackage{orcidlink}

\begin{document}

\title{Empowering Embodied Visual Tracking with Visual Foundation Models and Offline RL} 

\titlerunning{Empowering Embodied Visual Tracking with VFMs and Offline RL}

\author{Fangwei Zhong\inst{1, 2\footnotemark[1]}\orcidlink{0000-0002-0428-4552}\and
Kui Wu\inst{3\footnotemark[1]}\orcidlink{0009-0004-5567-2792}\and
Hai Ci\inst{4}\orcidlink{0000-0001-7170-277X} \and
Churan Wang \inst{1}\orcidlink{0000-0002-7699-0894} \and
Hao Chen \inst{5}\orcidlink{0000-0001-6816-5344}
}

\authorrunning{F.~Zhong et al.}

\institute{Peking University, Beijing, China \\
\and
State Key Laboratory of General Artificial Intelligence, BIGAI\&PKU,  Beijing, China
 \and
State Key Laboratory of Complex \& Critical Software Environment, Beihang University, Beijing, China\\
 \and
National University of Singapore, Singapore\\
 \and
City University of Macau, Macao, China\\
\email{\{zfw, cihai, churanwang\}@pku.edu.cn}
\email{\{wukui0099, sundaychenhao\}@gmail.com}
}
\maketitle
\renewcommand{\thefootnote}{\fnsymbol{footnote}} 
\footnotetext[1]{indicates equal contribution}
\renewcommand{\thefootnote}{\arabic{footnote}}
\input{Sec/0_abstract}   
\input{Sec/1_intro}

\input{Sec/2_related_work}

\input{Sec/3_method}
\input{Sec/4_experiments}

\input{Sec/5_conclusion}


%
%
\bibliographystyle{splncs04}
\bibliography{main}
\clearpage
\appendix
\input{Sec/6_appendix}
\end{document}

%% file: Sec/0_abstract.tex
\begin{abstract}
Embodied visual tracking is to follow a target object in dynamic 3D environments using an agent’s egocentric vision. 
This is a vital and challenging skill for embodied agents. However, existing methods suffer from inefficient training and poor generalization. In this paper, we propose a novel framework that combines visual foundation models (VFM) and offline reinforcement learning (offline RL) to empower embodied visual tracking. We use a pre-trained VFM, such as ``Tracking Anything”, to extract semantic segmentation masks with text prompts. We then train a recurrent policy network with offline RL, e.g., Conservative Q-Learning, to learn from the collected demonstrations without online interactions. To further improve the robustness and generalization of the policy network, we also introduce a mask re-targeting mechanism and a multi-level data collection strategy. In this way, we can train a robust policy within an hour on a consumer-level GPU, e.g., Nvidia RTX 3090. We evaluate our agent on several high-fidelity environments with challenging situations, such as distraction and occlusion. The results show that our agent outperforms state-of-the-art methods in terms of sample efficiency, robustness to distractors, and generalization to unseen scenarios and targets. We also demonstrate the transferability of the learned agent from virtual environments to a real-world robot. 
\footnote{Project Website: \url{https://sites.google.com/view/offline-evt}}
\end{abstract}

%% file: Sec/1_intro.tex
\section{Introduction}
\label{sec:intro}
Embodied agents, such as robots or virtual avatars, need to track their interested objects (target) in their surroundings with visual observation while performing social tasks. For example, a robot may need to follow a user’s path to prepare for help. Embodied visual tracking (EVT) has a wide range of applications, including drones~\cite{ci2023proactive}, mobile robots~\cite{wang2018accurate}, and unmanned vehicles~\cite{jin2022conquering, jin2023demrl}. 
However, EVT faces several challenges in applications:
1)\textbf{Training Efficiency}: Training the tracking policy by Reinforcement Learning (RL) is a common approach~\cite{luo2018end, zhong2018advat}, but it is computationally intensive and time-consuming. The agent requires extensive trial-and-error interactions with the environment for learning, often taking more than 12 hours.
2)\textbf{Cross-Domain Generalization}: Agents are required to track unseen targets with different visual appearances and motion patterns in diverse environments. Thus, it faces a large domain gap between the training and testing environments.
3)\textbf{Spatial and Temporal Reasoning}: When tracking in complex environments, the agent must infer the dynamics of the scene spatially and temporally to handle obstacles, occlusions, and distractors.
4)\textbf{Real-Time Interaction}: As the target moves, the embodied agent must react quickly with limited computational resources. This imposes challenges in balancing efficiency and accuracy at inference time.

\begin{figure*}[t]
\centering 
\includegraphics[width=\textwidth]{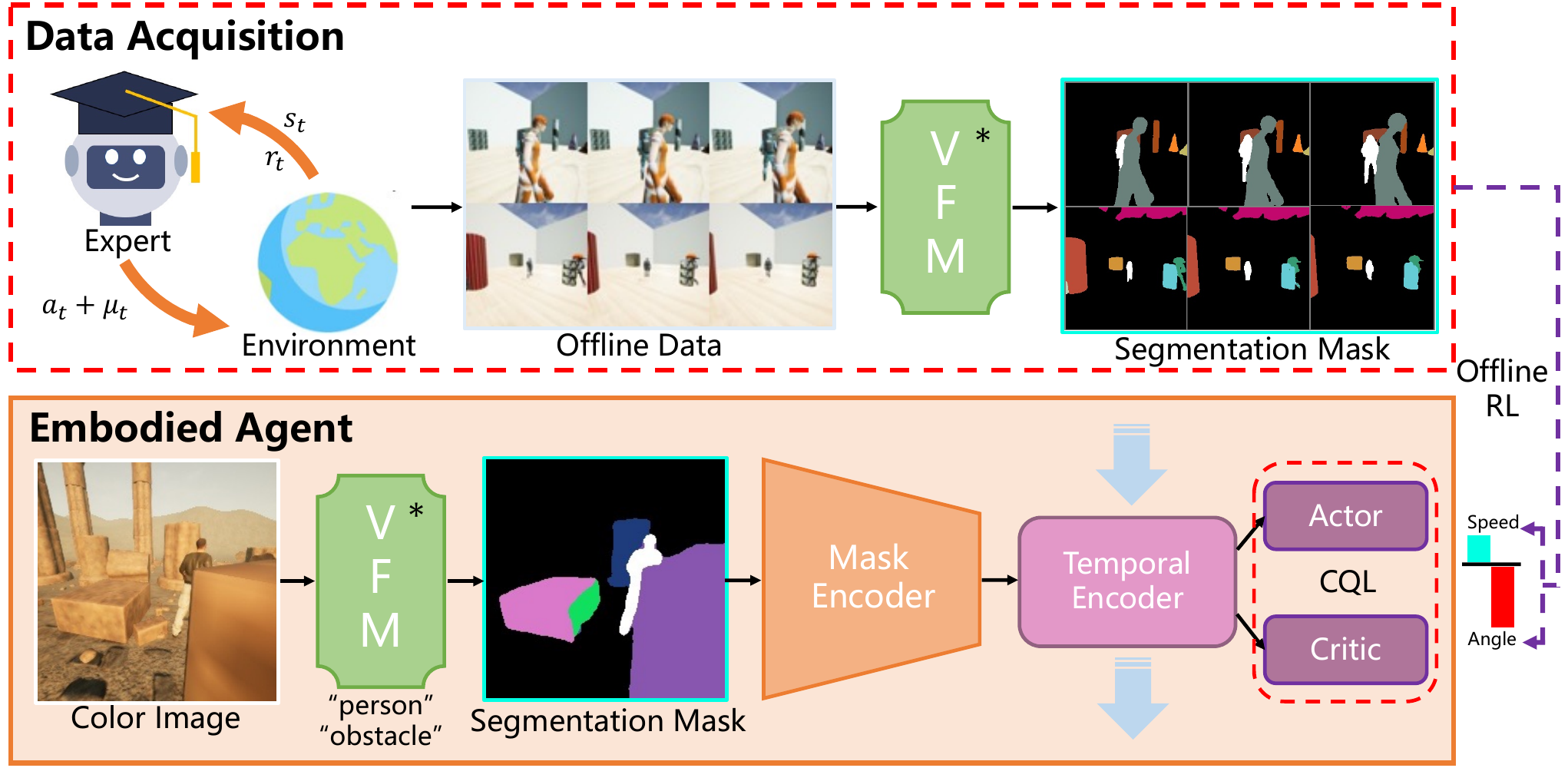} 
\caption{The overall framework of our proposed method. 
For data acquisition, a state-based policy is employed to collect diverse image-action-reward trajectories through interactions with various complex environments. We augment the policy by adding different levels of perturbation in actions. 
The observed color images are encoded into text-conditioned segmentation masks, which highlight the target object (white), show the obstacle (colorful), and remove the background noise (black). Subsequently, we employ offline reinforcement learning, such as conservative Q-learning (CQL), to train the recurrent policy network, which outputs actions based on the segmentation masks.
}
\label{Fig.1}
\end{figure*}

In this paper, we propose an efficient training framework for embodied visual tracking, by integrating visual foundation models and offline reinforcement learning. 
This framework stands out as an efficient method for training a visual policy. Our framework consists of four main components: 1) A data acquisition pipeline can generate multi-level imperfect state-based expert policy to automatically collect diverse trajectories in an augmentable virtual environment. 2) A text-conditioned segmentation mask, provided as an abstracted representation of the target and obstacles. 3) A recurrent policy network to filter the noise in the masks and output temporal consistent actions for tracking. 4) An offline reinforcement learning algorithm trains the policy network with the collected data.
Within this framework, we train a robust tracking policy within an hour on a consumer-level GPU, e.g., Nvidia RTX 3090. 
We evaluate our tracker on a set of high-fidelity virtual environments that simulate diverse and challenging situations, such as active distraction, occlusion, and reflection. 
The results show that our framework achieves superior performance in terms of accuracy, robustness, and generalization.
We also validate the sim2real transferability by deploying our method in a real-world robot.

Our main contributions are in threefolds:
1) We propose a framework that incorporates the visual foundation models and offline reinforcement learning with practical techniques, such as text-conditioned segmentation mask, re-targeting mechanism, and recurrent policy, to efficiently build a robust embodied visual tracking agent. 
2) We introduce a data acquisition method with multi-level policy perturbation and collect an offline dataset for offline RL.
3) We evaluate the proposed method in both virtual and real environments, showing its significant improvements in robustness and generalization.

%% file: Sec/2_related_work.tex
\section{Related work}
\label{sec:related}
\textbf{Embodied Visual Tracking (EVT)} is an active research area in embodied vision~\cite{luo2018end, luo2019pami, zhang2018coarse, xi2021anti, li2020pose, devo2021enhancing, zhong2019ad, zhong2018advat, zhong2021distractor, zhong2023rspt}. 
Luo, et al~\cite{luo2018end} first trained an end-to-end visual policy via RL and successfully deployed the policy trained in a simulator on a real-world robot~\cite{luo2019pami}.
Zhong, et al.~\cite{zhong2019ad, zhong2018advat} use multi-agent games to improve the robustness and generalization of the tracker in unseen environments.
Recently, Meta has adopted a similar end-to-end RL method for embodied social navigation tasks~\cite{puig2023habitat}. Zhong, et al.~\cite{zhong2021distractor} and Bajcsy, et al. ~\cite{bajcsy2023learning,chenALDI} use cross-modal teacher-student learning strategy with multi-agent games to train a robust visual tracking policy. However, all these methods are trained by online interactions, leading to a high cost in training time. In addition, training time might increase when transferring to complex real-world environments, such as obstacle occlusions, active distractors, and temporary targets.
To improve training efficiency, we first introduce an offline reinforcement learning paradigm for embodied visual tracking, making it feasible to efficiently train a robust tracker in an hour.

\textbf{Offline Reinforcement Learning (Offline RL)}~\cite{fujimoto2019off, kumar2020conservative} has been first proposed as a paradigm that learns optimal policies from fixed offline datasets, avoiding the need for online data collection, which aims to increase the training efficiency. Shah, et al.\cite{shah2022offline} presented the first offline RL system for robotic navigation, which uses previously collected data to optimize user-specified reward functions in the real world. Yet, this system depends on a fixed reward function, making it less adaptable to various environments. With superior training efficiency, Offline RL still faces several challenges, such as data distributional shifts, inaccurate state reward estimation, and exploration-exploitation trade-offs. 

\textbf{Visual Reinforcement Learning (Visual RL)} faces challenges due to the high-dimensional and noisy observation space. End-to-end methods typically encode input images directly into a latent space and often suffer from inefficiency and poor generalization. To address these issues, several works have explored the impact of mid-level representations on final performance~\cite{zhou2019does, mousavian2019visual, yang2018visual, sax2020learning, yuan2022pretrained,chenMPAD}, indicating that while mid-level representations generalize well across domains, the optimal representation varies by task. For instance, CSR~\cite{gadre2022continuous} extracting bounding boxes and masks to create an object state graph, which enhances spatial awareness of visual representation for embodied agent navigation tasks~\cite{chenLAGCN}. RSPT~\cite{zhong2023rspt} introduces a motion-aware structure representation for tracking in cluttered environments. In this paper, we leverage visual foundation models to generate text-conditioned object masks as state representations. The learned policy can generalize well to unseen environments and targets. Our solution may also inspire other visual reinforcement learning tasks.  

\textbf{Visual Foundation Models (VFMs)}\cite{awais2023foundational,chen2024macro} learn versatile visual representations from large-scale data for tasks like scene understanding, object detection, and segmentation. However, in embodied vision, where agents interact with dynamic environments, VFMs face challenges such as lack of spatial awareness and robustness. For instance, CLIP\cite{radford2021learning} struggles with precise localization, and SAM~\cite{kirillov2023segment} and Grounding-DINO~\cite{liu2023grounding} are not robust to noise, occlusion, and motion. To overcome these limitations, models like TAM~\cite{yang2023track}, E-SAM~\cite{xiong2023efficientsam}, DEVA~\cite{cheng2023tracking}, and SAM-Track~\cite{cheng2023segment} integrate VFMs with other models to improve segmentation and tracking.
In robotic tasks, VFMs are used for state representations or goal states, as seen in EmbCLIP~\cite{khandelwal2022simple} for navigation and DALL-E-Bot~\cite{dall-e-bot} for object rearrangement. LfVoid~\cite{gao2023pretrained} links instructions and visual goals in reinforcement learning. In this paper, we leverage text-based video object segmentation models to generate text-conditioned segmentation masks for embodied visual tracking, enhancing spatial and temporal understanding and improving policy generalization with a re-targeting mechanism.

%% file: Sec/3_method.tex
\section{Task Definition}
Embodied visual tracking is a typical embodied vision task that involves both visual perception and movement control. The tracking agent needs to cope with occlusions, distractors, and high-dynamic environments.
The task can be formulated as a Partial Observable Markov Decision Process (POMDP), with state $s \in \mathcal{S}$, partial observation $o \in \mathcal{O}$, action $a \in \mathcal{A}$, and reward $r \in \mathcal{R}$. At each time step $t$, the agent observes an image $o_t$ captured by its embodied vision system and outputs an action $a_t$ to maintain a relative position to the target object. The agent receives a reward $r(s_t)$ that reflects how well it tracks the target. In the agent-centric coordinate system, the reward function is defined as $    r = 1- \frac{\vert\rho - \rho^*\vert}{\rho_{max}} - \frac{\vert\theta-\theta^*\vert}{\theta_{max}}$,
where $(\rho, \theta)$ denotes the current target position relative to the tracker, $(\rho^*, \theta^*)=(2.5m, 0)$ represents the expected target position, i.e., the target should be $2.5m$ in front of the tracker. The error is normalized by the field of the view $(\rho_{max}, \theta_{max})$. This reward function is similar to the one used in \cite{zhong2019ad, zhong2023rspt}. The goal of the agent is to learn a policy $\pi(a_{t}|s_{t})$ that maximizes the expected cumulative reward over a finite horizon $T$, i.e., $E_{\pi} \left[\sum_{t=0}^T r_t\right]$. 

In this paper, we consider the offline training setting that aims to learn a policy with a fixed dataset of transitions $\mathcal{D}= \{ (\mathcal{T}_0, \mathcal{T}_1, ..., \mathcal{T}_T )_i | i = 1...N \}$, where $T$ is the episode length, $N$ is the number of episodes. A transition is a tuple of state, action, reward, and next state, denoted as $\mathcal{T}_t=(s_t, a_t, r_t, s_{t+1})$. In this setting, the agent is not required any online interaction with the environment during training. This is crucial because online interaction has been identified as a primary bottleneck in enhancing the efficiency of the training process. 
In this setting, the main challenge is to find an effective solution to improve the generalization of the policy to handle the unseen state transition that is not contained in the offline dataset.

\section{Method}

In this section, we introduce our framework to learn visual embodied tracking in an offline manner. First, we employ the visual foundation models, e.g., DEVA~\cite{cheng2023tracking} or SAM-Track~\cite{cheng2023segment}, to build a generalizable text-conditioned segmentation mask as state representation and a multi-level expert generation strategy to collect diverse demonstrations.  At the end, we train a recurrent policy network via offline reinforcement learning, e.g., Conservative Q-learning~\cite{kumar2020conservative}. 
\subsection{Text-Conditioned Segmentation Mask}
We find that vanilla end-to-end offline reinforcement learning does not work well for embodied visual tracking (EVT) due to the high-dimensional visual observation and the complex dynamic environment. We propose a text-conditioned segmentation mask as the state representation for EVT, which should be spatial-temporal consistent, domain-invariant, and efficient for real-time execution. We use an open-vocabulary segmentation model and a re-targeting mechanism to obtain a temporal-consistent text-conditioned segmentation mask. The open-vocabulary tracking model can abstract and generalize the state representation by combining semantic knowledge and visual features. This representation can be provided by SAM-Track~\cite{cheng2023segment}, DEVA~\cite{cheng2023tracking}, and other open-vocabulary segmentation models utilizing text prompts. The re-targeting mechanism can make the color of the target in the mask consistent across different episodes and enhance the contrast between the target and the environment, improving the robustness to visual distraction. We introduce the details of the two steps as follows:

\begin{figure}[t]
\centering
\includegraphics[width=1.0\linewidth]{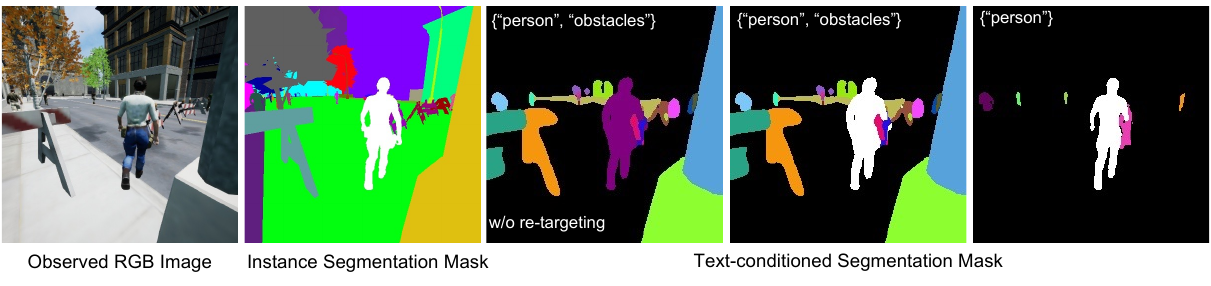}
 \caption{
 Examples of an observed RGB image and the corresponding instance segmentation mask (ISM) from UnrealCV~\cite{qiu2017unrealcv} and Text-conditioned segmentation mask from DEVA.
The middle one is the original mask from DEVA~\cite{cheng2023tracking}. 
 The two right masks utilize re-targeting to emphasize the target with different text prompts.
 }
\label{fig:mask}
\end{figure}

1) \textbf{Segmentation with Text Prompts}:
We use the text prompt to specify the category of the desired target object and the related background objects for segmentation. For example, when tracking a person in a cluttered environment, we can specify the prompt as $\left \{``person", ``obstacles" \right \}$, thereby drawing a segmentation mask identifying each person and obstacle uniquely. 
In contrast to \emph{Instance Segmentation Mask (ISM)}, \emph{Text-Conditioned Segmentation Mask} can effectively eliminate extraneous background details, thus reducing the presence of irrelevant noise and mitigating the domain gap. If we only segment the target, the obstacle will be missed in the representation, leading to collision in complex environments.
Specifically, at time step $t$, the VFM with text prompot $\phi _{prompt}$ extracts a segmentation mask $S_{t}$ from the image $O_{t}$, expressed as $S_{t}=\phi _{prompt}(O_{t})$. The mask contains $K$ 
instances, each denoted as $m{(i,c)}$, 
where $m(i, c) \in S_{t}$, $c$ is the object category $c \in \left \{``person", ``obstacles" \right \}$ and $i$ is the instance ID.
Figure~\ref{fig:mask} shows the ISM and text-conditioned masks in different configurations.

2) \textbf{Re-targeting Mechanism}: We can not directly identify the target from the original masks provided by the VFMs, as they randomly assign different colors to each instance. Such masks will require our policy to learn to reason the target from a sequence of masks. To explicitly identify the target in the mask, we introduce a re-targeting mechanism. At the initial frame of each episode, we specify the initial target mask using a scoring system that prioritizes saliency, based on size and centrality. Masks covering over 15\% of the image are scored with: $Score=s_{m_{(i,c)}}*(1-|e_{pos}|)$, where $s_{m_{(i,c)}}$ is the mask's area size and $e_{pos}$ is the normalized position error from the mask center to the image center. The mask with the highest score is selected as the target object $mask_{(i,c)}$, ensuring that it is large and centrally located. We record the index $i$ as the target ID. In the subsequent propagation procedure, all object masks associated with the target ID have their RGB values manually adjusted to $(255,255,255)$, resulting in a white color representation, as shown in Figure~\ref{Fig.1}. 


\subsection{Multi-level Demonstration Collection}
\label{sec:data acquisition}
For offline policy learning, the diversity of the demonstration determines the skill level of the learned policy. However, in many cases, it is unrealistic and inappropriate to collect training data by deploying a real robot because of the high cost of trial-and-error interactions. Besides, the accurate reward signal is necessary for learning, which is hard to obtain in real-world scenarios.
We use an augmented virtual environment to automatically collect large-scale tracking demonstrations for offline policy learning. By leveraging UnrealCV~\cite{qiu2017unrealcv}, we control environmental factors to enhance diversity and access ground truth states for building an expert for EVT. Our data collection process involves three steps:

1) \textbf{Environment Randomization:}
We stochastically initialize various environmental factors, including lighting conditions, the number of obstacles and players, as well as the positions, rotations, and textures of these elements. This configuration intentionally includes challenging scenarios, such as the presence of multiple distracting agents and unknown environmental structures. we randomly select a tracker and a target from the pool of all players. Subsequently, the expected location of the tracker is positioned $2.50 m$ behind the target. The tracker’s expected rotation angle is carefully adjusted to maintain the target at the center of the field of view, as shown in Figure~\ref{fig:mask}. 

2) \textbf{Multi-Level Data Generation}: 
We develop a navigation system in Unreal Engine to generate diverse trajectories for the target and distractors. The system enables agents to autonomously navigate to any reachable locations via the shortest path. For tracking, we use a state-based PID controller as the \textit{expert policy}. To simulate the policy with different skill levels, we add noise to the output actions. We define a probability threshold $\mu$ for switching to random actions for a random number of continuous steps. By adjusting $\mu$, we can generate multi-level demonstrations, from random walking to closed following. 
We apply the state-based tracking policy for 500 steps per episode, and record the tracker’s visual observations, rewards, actions, and done flags of each time step. 
Each data element is a tuple of ($o_t$, $a_t$, $r_t$, $done$), where $o_t$ is the observation (first-person view RGB image of the tracker), as shown in Figure~\ref{fig:mask}, $a_t$ is the action taken by the agent, $r_t$ is the reward of the current step, and done is a boolean indicating the end of the episode. The data elements are ordered by their time stamps.

3) \textbf{Pre-Processing Raw Observation:}
Before training, we use the visual foundation model and re-targeting mechanism to transform all raw-pixel images to text-conditioned segmentation masks, as shown in Figure~\ref{fig:mask}.
Notably, the re-targeting mechanism is a simple but effective method to highlight the target from other objects in the segmentation mask (the target is shown in white).

\subsection{Learning Recurrent Policy via Offline RL}

The recurrent policy network contains a segmentation mask encoder, a temporal encoder, and an actor-critic network. We train it using offline reinforcement learning with the collected dataset, as illustrated in Figure~\ref{Fig.1}.

\textbf{Recurrent Policy Networks:}
Drawing from prior experience, the agent may encounter challenges related to partial observability and non-Markovian movements~\cite{zhong2021distractor}. For instance, abrupt changes in speed or direction near obstacles could lead to the loss of the target within the field of view. Traditional single-frame state representations lack temporal information and are ill-equipped to handle such scenarios during long-horizon interactions with the environment. To address these issues, we adopt a standard CNN-LSTM architecture as our recurrent state encoder, which effectively captures long-term hidden state representations and also maintains a real-time inference. In offline training, we first sample a batch sequence of data from the collected offline dataset as input, each sequence of data is expressed as $\{ (S_{i}, a_{i}, S_{i}^{'}, r_{i})...(S_{i+L}, a_{i+L}, S_{i+L}^{'}, r_{i+L}) \}$
where $i$ is a randomly selected starting frame in an offline dataset. $S_{i}^{'}$ is the next state of $S_{i}$. $L$ is the predefined sequence length. In the training phase, the sequence of the pre-processed segmentation masks $S_{i}, S_{i+1}...S_{i+L}$ are encoded into state representation $x_{i},x_{i+1}...x_{t}$, which is used in Q network to learn offline action policy $\pi_{off}$. In the testing phase, the current segmentation result $S_{t}$ and the last hidden state $h_{t-1}, c_{t-1}$ are used to get the action $a_{t}$. 

\textbf{Conservative Q-learning with SAC:} To optimize the policy network, we employ the Conservative Q-learning (CQL)\cite{kumar2020conservative} with the Soft Actor-Critic (SAC) networks\cite{haarnoja2018soft} for our methods, referred to as CQL-SAC. CQL-SAC consists of three neural networks: two critic network $Q^{1}_\theta$, $Q^{2}_\theta$, used for estimate Q values and an actor network $V_{\phi}$, used for learning control policy. $\theta$ and ${\phi}$ are network parameters. Moreover, CQL-SAC uses an entropy regularization coefficient $\alpha$, which controls the degree of exploration. The Q-functions are updated by minimizing the following objective function:

\begin{equation}
\begin{aligned}
&L_\theta(D)=
\mathbb{E}_{\mathbf{s} \sim \mathcal{D}}\left[\log \sum_{\mathbf{a}} \exp (Q^i_\theta(\mathbf{s}, \mathbf{a}))-\mathbb{E}_{\mathbf{a} \sim \pi_\phi(\mathbf{a} \mid \mathbf{s})}[Q^i_\theta(\mathbf{s}, \mathbf{a})]\right] \\
& +\frac{1}{2} \mathbb{E}_{\mathbf{s}, \mathbf{a}, \mathbf{s}^{\prime} \sim \mathcal{D}}\left[\left(Q^i_\theta(\mathbf{s}, \mathbf{a})- \left(r+\gamma E_{a^{\prime} \sim \pi_\phi}\left[Q_{min}\left(s^{\prime}, a^{\prime}\right)-\alpha \log \pi_\phi\left(a^{\prime} \mid s^{\prime}\right)\right]\right)\right)^{2}\right]
\end{aligned}
\end{equation}
where $i \in \{1,2\}$, $Q_{min}=min_{i \in \{1,2\}}Q_\theta^i$, $\gamma$ is the discount factor, $\pi_\phi$ is the policy that derived from the actor network $V_\phi$. The first term is the CQL regularization term, which encourages the Q-function to be close to the log-density of the data distribution, and the second term is the standard SAC objective.

%% file: Sec/4_experiments.tex
\section{Experiments}

In this section, we conduct comprehensive experiments to validate: 1) the training efficiency, 2) the generalization ability to unseen environments and unseen target categories, 3) the robustness to visual distractions, 4) the effectiveness of each contributed module, and 5) the transferability to real-world scenarios.

\subsection{Experimental Setup}
\subsubsection{Environments:}
For training, we employ an augmentable virtual environment called \emph{Complex Room}. This environment includes a variety of obstacles, distractors, textures, and lighting conditions, providing intricate and diverse scenarios. To further enhance data diversity, we introduce randomness in the size, shape, textures, and lighting of all objects in each episode. The baseline methods of those using online RL are trained in this environment. We also collect the demonstration for offline policy learning, e.g., offline RL or Behavior Cloning.

For testing, we evaluate the agent in five testing environments: 1) \emph{Simple Room}, serving as a base environment to verify the model's basic tracking ability. 2) \emph{Parking Lot}, characterized by obstructed and dim lighting conditions. 3) \emph{Urban City}, a typical urban street scene with mirror-like reflections on the ground. 4) \emph{Urban Road}, similar to Urban City but with lower obstacles. 5) \emph{Snow Village}, characterized by rough terrain and complex backlight lighting conditions.
To evaluate the robustness of the trackers, we introduce varying numbers of distractors into the environment, denoted by the letter ``D”. For example, ``Parking Lot (4D)” indicates an environment with four distractors.
To evaluate the generalization on unseen targets, we add four kinds of animal in \emph{Simple Room}, including horse, sheep, dog, and pig.

\subsubsection{Evaluation:}

In our evaluation setting, we follow the evaluation setup in previous works~\cite{zhong2023rspt}, which sets the maximum step length in an episode to 500 steps. We define the visible region of the tracker as a 90-degree fan-shaped sector with a radius of 750 cm. A tracker succeeds if it keeps the target in sight for the whole episode (500 steps). The tracker fails if the target continuously leaves this region for more than 50 steps. In each episode, the agent terminates with a success or failure state, and we reset the environment and the agent for the next episode. With the above settings, we employ the following three evaluation metrics to analyze the tracking performance: 1) \emph{Accumulated Reward~(AR)}. We assess our model in each environment over 100 episodes and compute the average accumulated reward from these episodes to gauge the tracking performance's accuracy. 2) \emph{Episode Length~(EL)}. We determine the average episode duration over 100 episodes, using termination rules that mirror long-term tracking performance.
3) \emph{Success Rate~(SR)}. We calculate the average success rate across 100 evaluation episodes to assess the model's robustness.

\subsubsection{Baselines:}

We conduct a comprehensive comparison of our proposed approach against eight state-of-the-art models. Specifically, we compare our methods with: 
1) \emph{DiMP}~\cite{bhat2019learning}: This model uses a pre-trained video tracker to provide a target bounding box as scene representation and tune a PID controller to control the tracker.
2) \emph{SARL}~\cite{luo2018end, luo2019pami}: This online RL method encodes RGB input into hidden visual features and trains an end-to-end Conv-LSTM network via reinforcement learning.  
3) \emph{AD-VAT}~\cite{zhong2018advat}: This is an asymmetric dueling mechanism that trains the RL-based tracker with a learnable adversarial target to improve the robustness of the end-to-end tracker.
4) \emph{AD-VAT+}~\cite{zhong2019ad}: 
The plus version further applies a two-stage training strategy for learning in complex environments with obstacles.
5) \emph{RSPT}~\cite{zhong2023rspt}: RSPT is the state-of-the-art RGB-D tracker that additionally uses Depth images to construct a structure-aware representation. It combines an off-the-shelf video tracker, mapping module, and trajectory predictor to construct the representation and train the policy via RL under the AD-VAT mechanism.
6) \emph{Behavior Cloning~(BC)}: This is a classic offline policy learning method, which is widely adopted in robotics and autonomous driving. The policy is supervised by expert actions.
7) \emph{Cross-modal Teacher-Student Learning (TS)}\cite{zhong2021distractor}: This is an extension of the BC method by using a pose-based teacher to provide online suggestions for the vision-based student while interacting with environments. Note that the policies for the teacher, target, and distractors are trained in a multi-agent game.
8)\emph{Dagger}:  This is a classic algorithm designed to iteratively improve policy learning by aggregating datasets of expert actions and the learner's actions, thereby mitigating the compounding error of the initial imperfect policy. To be fair, we use our proposed text-conditioned segmentation mask as input, the same expert policy as the BC method, and the same training environment.
To ensure a fair comparison, we maintain the original parameter settings for all baseline methods. 

We also integrate eight configurations for systematic ablation, detailed in Table \ref{tab:config}. The first six configurations result from combining various VFM models while omitting specific ablation components. Instance Segmentation Mask (ISM) and RGB data are directly extracted from the game engine (Figure \ref{fig:mask}). Additionally, we obtained the target bounding box coordinates as visual representations. 
For each configuration, we use the same offline dataset and adhere to predefined settings for training and evaluation.

\subsubsection{Implementation Details:} \label{implementation} In our experiments, we used a continuous action space for agents. The action space contains two variables: the angular velocity and the linear velocity. 
Angular velocity varies between $-30^{\circ}/s$ and $30^{\circ}/s$, 
while linear velocity ranges from $-1\ m/s$ to $1\ m/s$.
To train our model, we use an Adam optimizer with a learning rate of 3e-5 and update the LSTM network every 20 steps. We set the mini-batch size to 32. The whole training process is run on one Nvidia RTX 3090 GPU. 
For offline policy learning, we collect demonstrations on the \emph{Complex Room} with 8 distractors.
We employ an imperfect expert policy with multi-level action noise (as described in Section~\ref{sec:data acquisition}) to generate an offline dataset with 42k steps of data as our training dataset for offline RL training, which we referred to as \emph{multi-level imperfect expert dataset}. Behavior Cloning (BC) is another offline RL method, which requires perfect expert demonstrations for training. Therefore, we adopt the same procedure but remove all random factors to collect 42k steps of \emph{expert dataset} for BC.

\subsection{Training Time-performance Superiority}
The first column of Table~\ref{tab:main_results} indicates the detailed training time of each method.
All online baselines (SARL, AD-VAT, AD-VAT+, RSPT, TS, Dagger) take at least 12 hours of training. SARL, AD-VAT+, TS, and Dagger take more than 24 hours to train a comparable tracker. Our proposed method takes 1 hour to outperform all other methods, which shows the advantage of our proposed method in training efficiency. 
DiMP uses a pre-trained video tracker with a PID controller, which does not need additional training time. However, it needs to manually tune the PID parameters and cannot handle complex environments.

\vspace{-0.3cm}
\subsection{Generalization to Unseen Environments}
\vspace{-0.2cm}
We evaluate the tracking performance in 5 different unseen environments and show the results in Table~\ref{tab:main_results}. Our method, in conjunction with most baseline approaches, exhibits strong performance within the reference environments (\textit{Simple Room}), except the BC method. We attribute this discrepancy to the inherent limitation of the BC method in addressing data shift challenges during offline training. Although Dagger shows comparable long-horizon tracking ability in SimpleRoom, it only achieved a score of 148 in AR, which is lower than other methods. We attribute this to our simple expert policy, which does not provide perfect supervision for learning smooth dynamic control. Consequently, the tracker can keep the target in view, but struggles to control the distance perfectly. These shortcomings are exacerbated in irregular terrains like Snow Village, where the tracker can easily get lost. In unseen environments, our method shows superior long-horizon tracking ability, perfectly following the target in Urban City. It consistently outperformed all baselines except the RSPT method (which uses depth images) in the Urban Road environment. The success rate of our method is only marginally lower by $0.04$ in Urban Road. The slight advantage of the RSPT method in Urban Road is due to its depth-based structure-aware motion representation, crucial for navigating low-height obstacles. In environments with complex light conditions and irregular terrains, our approach exceeds RSPT, demonstrating superior performance and robustness to visual distractions. 

\begin{table}[!t]
    \centering
    \caption{Quantitative results compared with baselines in unseen environments. From left to right, the three numbers of each cell represent the Average Accumulated Reward (AR), Average Episode Length (EL), and Success Rate (SR).}
    \resizebox{\textwidth}{!}{
        \begin{tabular}{l|ccccc|c}
        \hline
        Methods/training time & SimpleRoom & Parking Lot & UrbanCity & UrbanRoad & Snow Village & Mean\\ \hline
        DiMP / 0 hours  & 336/\textbf{500}/\textbf{1.00} & 166/327/0.48 & 239/401/0.66 & 168/308/0.33 & 110/301/0.43 & 204/367/0.58 \\ 
        SARL / 24 hours & 368/\textbf{500}/\textbf{1.00} & 92/301/0.22 & 331/471/0.86 & 207/378/0.48 & 203/318/0.31 &240/394/0.57\\       
        AD-VAT / 12 hours &  356/\textbf{500}/\textbf{1.00}    &  86/302/0.20 & 335/484/0.88 & 246/429/0.60 & 169/364/0.44 &238/416/0.62 \\ 
        AD-VAT+/ 24 hours & 373/\textbf{500}/\textbf{1.00}  &267/439/0.60 & \textbf{389}/\underline{497}/\underline{0.94} &  \underline{326}/471/0.80 & 182/365/0.44& 307/454/0.76\\ 
        TS / 26 hours & \textbf{412}/\textbf{500}/\textbf{1.00} & 265/472/\underline{0.89} & \underline{341}/496/\underline{0.94} & 308/480/0.84 & \underline{234}/\underline{424}/0.63 & \underline{312}/474/0.86 \\  
        RSPT / 12 hours  & \underline{398}/\textbf{500}/\textbf{1.00} & \textbf{314}/\underline{480}/0.80 & \underline{341}/\textbf{500}/\textbf{1.00} & \textbf{346}/\textbf{500}/\textbf{1.00} & \textbf{248}/410/\underline{0.80} & \textbf{329}/\underline{478}/\underline{0.92} \\
        Dagger / 24 hours & 148/498/0.98 & 115/446/0.78 & 133/448/0.78 & 102/396/0.60 & 47/154/0.04 & 109/388/0.64 \\ \hline
        BC / 1 hours  & 110/348/0.33 & 110/350/0.36 & 182/371/0.47 & 153/348/0.17 & 39/279/0.27 &119/339/0.32\\ \hline
        Ours/ 1 hours & 374/\textbf{500}/\textbf{1.00} & \underline{274}/\textbf{484}/\textbf{0.92} & 306/\textbf{500}/\textbf{1.00} & 300/\underline{496}/\underline{0.96}& 229/\textbf{471}/\textbf{0.87} & 296/\textbf{490}/\textbf{0.95}\\ \hline
        \end{tabular}
    }
    \label{tab:main_results}
\end{table}

\noindent
\begin{minipage}[l]{0.48\textwidth}
    \centering
    \captionof{table}{Evaluating the distraction robustness in the environment with distractors. (4D) represent there are 4 distractors in the environment. 
    }
    \resizebox{1.0\linewidth}{!}{
    \begin{tabular}{l|ccc}
    \hline
        Methods 
        & Parking Lot (2D) & UrbanCity (4D) & ComplexRoom (4D) \\ \hline
        DiMP 
        & 111/271/0.24 & 170/348/0.32 & 97/307/0.26 \\ 
        SARL 
        & 53/237/0.12 & 74/221/0.16 & 22/263/0.15 \\ 
        AD-VAT 
        & 43/232/0.13 & 32/204/0.06 & 16/223/0.16 \\ 
        AD-VAT+ 
        & 35/166/0.08 & 89/245/0.11 & 35/262/0.18 \\ 
        TS 
        & \underline{186}/\underline{331}/\underline{0.39}  & \underline{227}/\underline{381}/\underline{0.51} & \underline{250}/\underline{401}/\underline{0.54} \\
        BC 
        & 26/245/0.09 & -25/181/0.02 & 86/336/0.24  \\ 
        \hline
        Ours 
        & \textbf{192}/\textbf{425}/\textbf{0.63} & \textbf{272}/\textbf{472}/\textbf{0.92} & \textbf{354}/\textbf{479}/\textbf{0.88} \\ \hline
    \end{tabular}
    }
    \label{tab:robust}

\end{minipage}
\begin{minipage}[r]{0.48\textwidth}
		\centering
            \captionof{table}{The configuration of different settings in state representation. They all are trained via offline RL and tested on the \textit{Complex Room}.}
            \vspace{-0.1cm}
		\scalebox{0.48}{
		\begin{tabular}[t]{l|c|c|c|c|c|c}
                \hline
                & VFM & Seg & Text & Re-targeting & RNN & AR/EL/SR\\ \hline
                Ours & DEVA & \checkmark & \checkmark & \checkmark & \checkmark & 306/479/0.92\\ 
                Ours* & SAM-Track & \checkmark & \checkmark & \checkmark & \checkmark  &339/497/0.97\\ 
                Ours w/o RNN & DEVA & \checkmark & \checkmark & \checkmark & &192/413/0.60 \\
                Ours w/o retargeting & DEVA & \checkmark  & \checkmark & & \checkmark &131/394/0.57 \\
                DINOv2 & DINOV2 & \checkmark & & & \checkmark & 234/434/0.74  \\
                E-SAM & E-SAM & \checkmark & & & \checkmark & -34/253/0.19\\ \hline
                ISM & - & \checkmark & & & \checkmark & 8/300/0.23 \\
                RGB & - & & & & \checkmark & -63/206/0.03\\
                Bounding box & - & & & & \checkmark & -82/138/0.00 \\ \hline
        \end{tabular}
	}
     		\label{tab:config}
\end{minipage}

\begin{figure}[t]
\begin{minipage}[h]{1.0\linewidth}
    \centering
    \includegraphics[width=.6\linewidth]{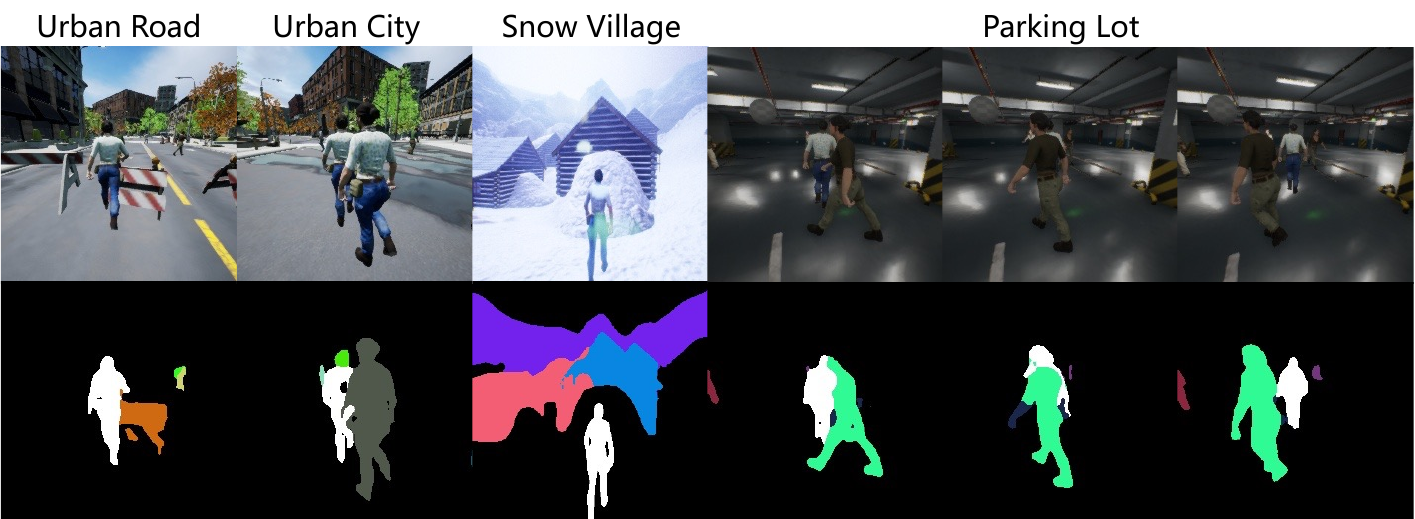}
    \includegraphics[scale=0.25]{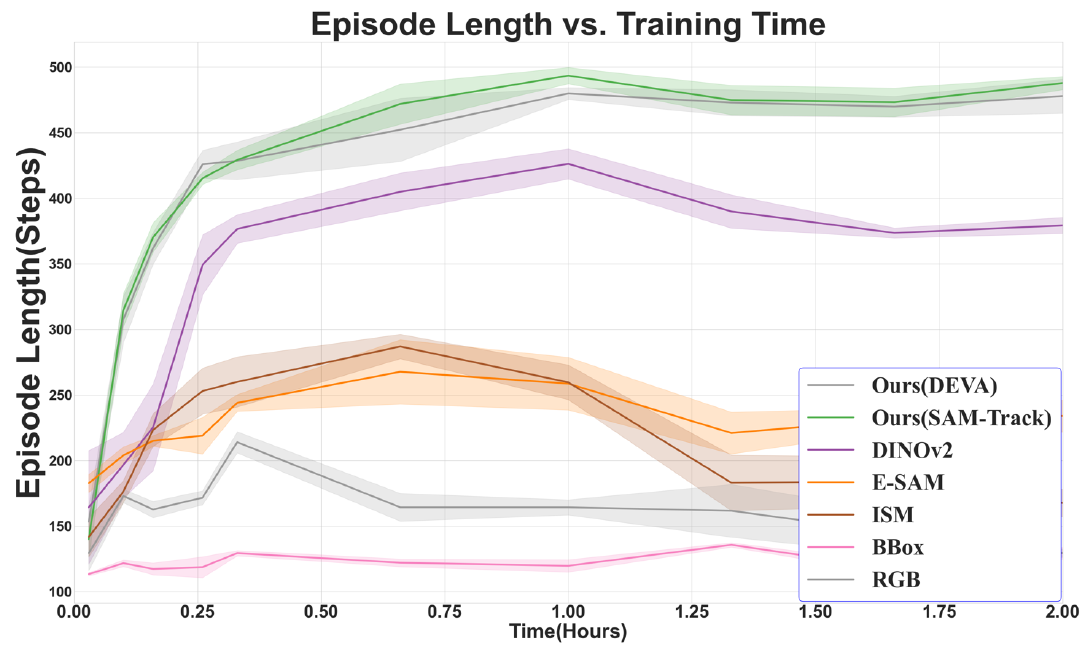}
\end{minipage}
\caption{ \textbf{Left:} The exemplars of the visual observation and segmentation mask provided by the vision foundation model in different testing environments. The three snapshots at \textit{Urban Road, Urban City, and Snow Village} show that the retargeting mechanism can distinguish the target, obstacles, and distractors with different colors in the mask. The sequence at (\textit{Parking Lot}) shows that the VFM can consistently identify the target even when the distractor fully occludes the target. \textbf{Right:} The learning curve of different state representations for offline RL validated in \textit{Complex Room} includes the mean and standard deviation of results obtained from training with 3 seeds.}
\label{fig:vis_abla}
\end{figure}
\subsection{Generalization to Unseen-category Targets}
We evaluate the generalization of our learned policy by tracking different target categories with distractors. 
We test the trackers in various conditions, including tracking different types of animals in the environment with four distractor animals, denoted as Animals(4D). At each episode, we randomly sample an animal as the target. We report the performance on Person (4D), where the target and distractors are persons in the SimpleRoom, as the reference. In each setting, the text prompt for VFM is the corresponding category of the target. 

The results in Table.~\ref{tab:unseen} show that our proposed method achieved tracking performance similar to that of the reference methods for single-category animals, and even outperformed them in some cases. However, when distractors were present, all three results slightly decreased. This indicates that our model successfully reduced the visual feature gap between categories, but the robustness of dynamic visual distraction across domains can still be improved.

\begin{table*}[!t]
    \centering
    \caption{Evaluating the generalization on the unseen category of the target in \textit{Simple Room}. 
     We directly adopt the agent on the unseen animals: horse, dog, sheep, and pig.
    }
        \resizebox{\linewidth}{!}{
    \begin{threeparttable}
    \begin{tabular}{l|ccccccc}
    \hline
         & Person (4D) & Animals (4D) & Horse & Dog & Sheep &  Pig \\ \hline
        Ours & 285/476/0.90 & 237/457/0.84 &254/500/1.00 &231/469/0.90 & 243/471/0.93 & 246/472/0.94 \\ \hline
        Ours w/o RNN & 177/454/0.81 & 207/425/0.77& 220/485/0.96 & 268/465/0.90 & 251/460/0.92 & 249/465/0.93 & \\ \hline
        Ours w/o Retargeting &93/343/0.37 & 27/291/0.25 &104/395/0.57. &54/319/0.33 & 32/292/0.30 & 33/360/0.30 \\ \hline
        DINOv2 &-53/236/0.05 & 8/237/0.20 &137/425/0.71 &-22/220/0.07 & -31/199/0.03 & -2/235/0.09 \\ \hline 
    \end{tabular}
    \end{threeparttable}}
    \label{tab:unseen}
\end{table*}

\subsection{Robustness to Visual Distraction}
We evaluate the robustness of our proposed method in the presence of visual distractions. Such distractors may occlude the tracker or present a texture similar to the target, thereby inducing visual ambiguity. 
The results in Table~\ref{tab:robust} show that distractors have a certain degree of impact on the tracker in all environments, but the overall Episode Length of our model remains above $400$. In addition, we can see that the success rate of our method significantly outperforms the state-of-the-art distraction-robust method~\cite{zhong2021distractor} in this setting. This demonstrates that our model has strong anti-distraction abilities and excellent robustness. We also provide an example sequence of our method in case of distraction in the \textit{Parking Lot} environments, as shown in Figure~\ref{fig:vis_abla}.

\subsection{Ablation Study}
\label{sec:ablation}

We conduct ablation studies to analyze the effects of each module in our method. 

\textbf{Dataset Size and Noise Level:}
We assess the impact of the size of the offline dataset by comparing models trained with $20k$ and $80k$ steps. The result illustrates that limited dataset size impedes the performance of policy while incrementing the dataset beyond a threshold yields marginal performance improvement. For behavior cloning, we extend the dataset size to 200k, 500k, and 1000k steps, observing limited improvement. We also explore the impact of noise level in our approach, which employs a constrained policy to estimate Q values. Pure expert data may lead to Q-function overestimation, and excessive noise can compromise policy quality. Three datasets are collected with varying noise levels, and optimal performance is achieved by noise level 2, cautioning against extreme noise or relying solely on expert data. The implementation details and results are shown in the supplementary material. 

\textbf{Text-conditioned Segmentation Mask:} We also analysis the effectiveness of the state representation for offline RL in EVT. We categorize state representations into two groups: those with VFMs and those without. The inference times vary by integrating different models, implementation details can be found in the supplementary material. Our ablation includes DEVA, SAM-Track, E-SAM, and DINOV2, highlighting the benefits of VFMs when combined with text prompts and re-targeting mechanisms.
The learning curves are shown in Figure \ref{fig:vis_abla}, confirming that VFMs with text guidance outperform others, as they effectively reduce background noise and enhance target identification. In particular, DINOv2's temporal consistency leads to its superior performance over E-SAM. The generalization of our framework is validated across different VFMs (DEVA and SAM-Track), indicating its adaptability to future advancements. Instance segmentation masks or RGB images fall short because of their high-dimensional nature and noise. In contrast, text-conditioned masks bridge the domain gap, improving performance. This study highlights the importance of text-conditioned masks in EVT, enhancing the robustness and effectiveness of our framework. 
\begin{figure}[t]
    \centering
    \includegraphics[width=\linewidth]
    {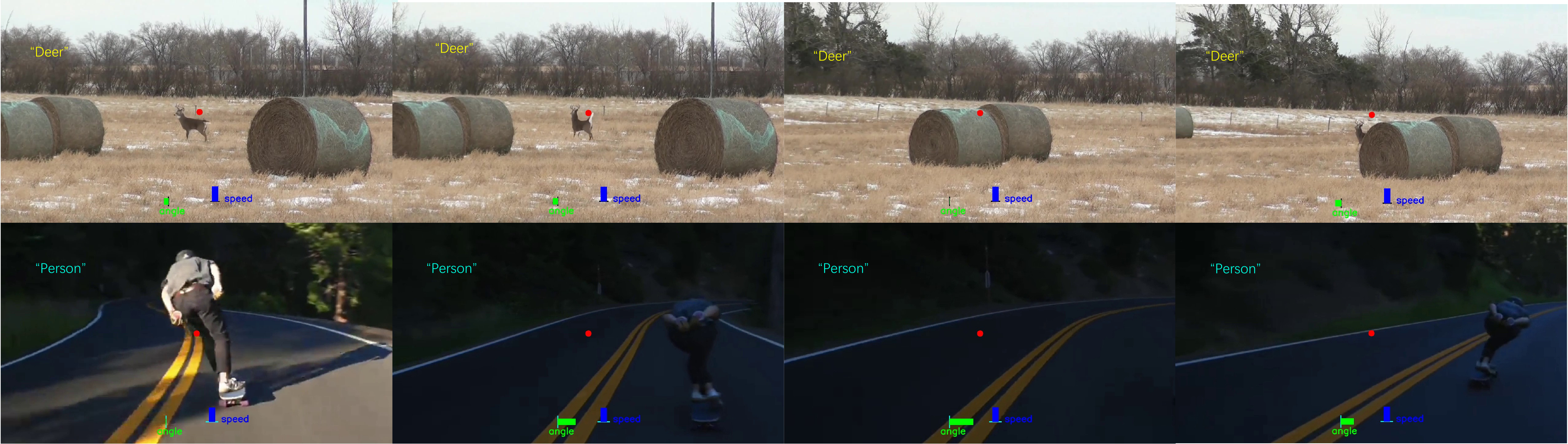}
    \caption{The output actions of our agent in two video sequences from VOT Challenge~\cite{VOT_TPAMI}. We separately use ``Deer" and ``Person" as prompts. The central red point signifies the image's center. The bottom green rectangle's width indicates angular velocity control from -30°/s to 30°/s, while the bottom blue rectangle's height indicates linear velocity control from -1 m/s to 1 m/s.}
    \label{fig:real world}
\end{figure}

\begin{figure}[!t]
\centering
\includegraphics[width=1\linewidth]{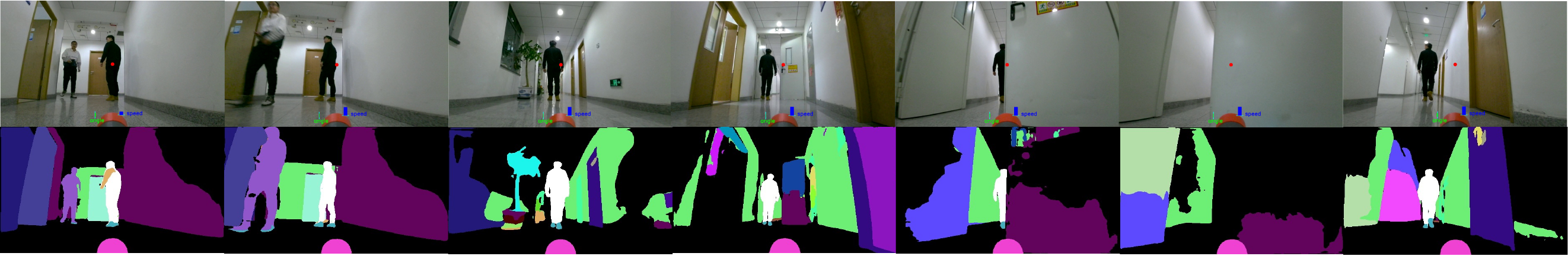}
\caption{The agent is deployed on a real robot, following a human in a complex corridor, navigating corners and hallways, involving pedestrians and a target occlusion.}
\label{figs:realworldVis}
\end{figure}
\textbf{Recurrent Policy:}
The RNN in the recurrent policy is also crucial, as the result shown in Table \ref{tab:unseen}. 
We trained an alternative model without the temporal encoder (w/o RNN). The results indicate a noticeable decline in performance in Person(4D) and Animals(4D), highlighting the effectiveness of temporal information in handling visual distractions.
\subsection{Transferring to Real-world Scenarios}

We further validate the cross-domain generalization of the learned tracker in real-world videos from the VOT dataset~\cite{VOT_TPAMI} and a mobile robot. 
We analyze the tracker’s performance by inputting video frames sequentially and comparing predicted actions to human intuition.
As shown in Figure~\ref{fig:real world}, the output actions align with our intuition, even when the target is occluded or out-of-view. For instance, in the top sequence, the tracker maintains near-zero angles and high speed during occlusion, then resumes left turns when the deer reappears. We also deploy our tracker on a mobile robot in an indoor hallway environment, featuring random pedestrian and obstacle distractions. We also present long continuous image sequences to show the robust tracking behavior of the robot controlled by our agent in a real-world corridor with human distractions and static obstacle occlusions, as is shown in Figure ~\ref{figs:realworldVis}. The implementation details and more real-world demos can be found in the supplementary material.

%% file: Sec/5_conclusion.tex
\vspace{-0.3cm}
\section{Conclusion}
 \vspace{-0.3cm}
\label{sec:con}

In this paper, we introduce a novel framework that marries visual foundation models with offline reinforcement learning, leading to an embodied visual tracking agent that is both training-efficient and robust.
Our empirical evaluations, conducted across various high-fidelity virtual environments, have demonstrated the framework’s exceptional ability to generalize. It has consistently outperformed existing methods, particularly in scenarios involving previously unseen targets and conditions. These advancements have opened new avenues for exploration and set a benchmark for future work in the domain. Future work will focus on real-world applicability, enhancing adaptability, and minimizing dataset dependence to fully harness the capabilities of embodied agents.

\section*{Acknowledgements}
This work was supported by the National Science and Technology Major Project (MOST-2022ZD0114900) and China National Post-doctoral Program for Innovative Talents (No. BX2021008).

%% file: Sec/6_appendix.tex
\section{Virtual-to-Real Transfer}

We apply our embodied vision tracker, trained with synthetic data, on a real-world mobile robot and test it in a complex indoor environment, including occlusions, obstacles, and active distractors, to verify the generalization of the learned policy.

\vspace{-0.2cm}
\subsection{Robot Setup}
\label{robot_setup}
\begin{figure}[h]
\centering
\vspace{-0.4cm}
\includegraphics[width=0.7\linewidth]{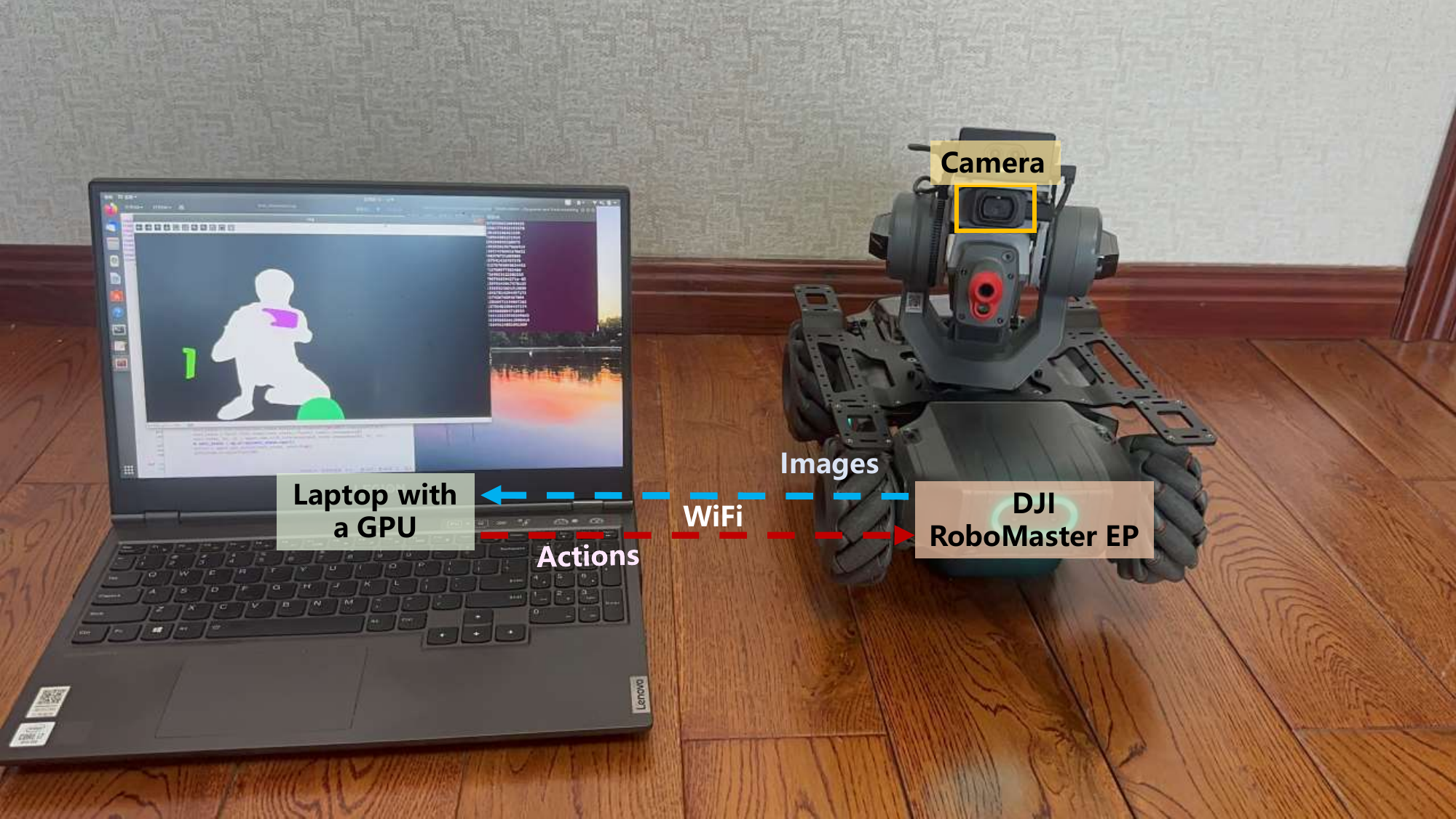}
\caption{The robotic deployment in a real-world scenario. The robot’s onboard camera captures images, which are then transmitted wirelessly to the laptop for real-time processing. The model on the laptop processes these images and outputs the actions. These actions are relayed back to the robot via WiFi, prompting it to adjust the movement. }
\vspace{-0.2cm}
\label{figs:real-deploy}
\end{figure}
\begin{table}[t]
        \centering
        \caption{We directly map our adopted action space(continuous actions) from virtual to real. The second and the third columns are the value ranges of velocities in the virtual and the real robot, respectively}
        \begin{tabular}{c|cc}
        \hline
        \multirow{2}*{Bound of Action} & Virtual_Linear(cm/step) & Real_Linear(m/s) \\
        ~ &Virtual_Angular(degree/s) & Real_Angular(rad/s) \\ \hline
        High & 100, 30 & 0.5, 1.0\\
        Low    & -100, -30 & -0.5, -1.0\\
        \hline
        \end{tabular}

        \label{tab:action_mapping}
    \end{table}
\vspace{-0.2cm}


We use the RoboMaster EP\footnote{https://www.dji-robomaster.com/robomaster-ep.html}, a 4-wheeled robot manufactured by DJI, as our experimental platform (Figure~\ref{figs:real-deploy}). Equipped with an RGB camera, the RoboMaster allows us to capture the image and provide direct control signals for both linear speed and angular motion via Python API. Leveraging its advanced design, our experiments benefit from precise control over the robot's movements. To facilitate real-time image processing, we rely on a wireless LAN connection to transmit data captured by the robot's camera to our laptop. Our laptop, equipped with an Nvidia RTX A3000 GPU, serves as the computational hub for running the model, enabling action prediction based on the received raw-pixel images. Subsequently, the corresponding control signals are generated and transmitted back to the robot, completing the closed-loop control system. This seamless integration of hardware and software components enables us to execute complex tasks with efficiency and accuracy, pushing the boundaries of sim-to-real experimentation. Note that, we can also deploy the embedded AI computing device for on-board computing, such as NVIDIA Jetson TX2, to replace the wireless solution.

In our training phase, we employ a continuous action space, allowing a mapping to real robot control commands. The details regarding this mapping are provided in Table \ref{tab:action_mapping}.

\subsection{Results on Real-World Environments}
\label{real-world demo}

For real-world deployment, in addition to the long image sequences shown in the main paper, we also evaluate our tracker in several different environments, featuring low-quality light conditions, wild animals (cat), and static obstacle occlusions. The videos are shown on: \url{https://sites.google.com/view/offline-evt}.

\section{Virtual Environments}
\begin{figure}[t]
\includegraphics[width=1\linewidth]{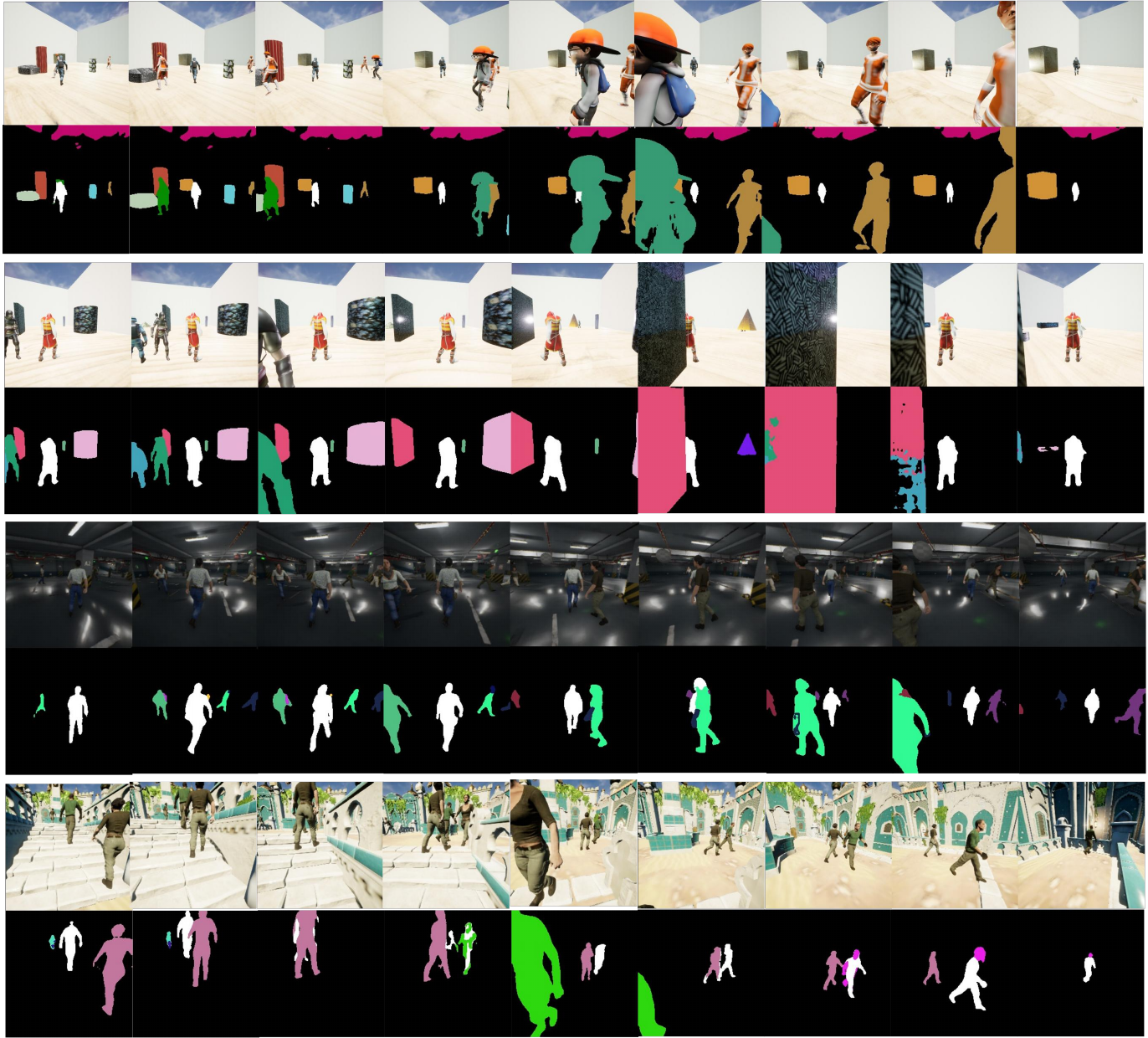}
\caption{The tracking demonstration of our agent in Virtual environments. The color image is the first-person view observation of the embodied agent. The mask on the bottom of the color image is the corresponding text-conditioned segmentation mask based on DEVA. }
\label{figs:SimVis}
\end{figure}

\begin{figure}[t]
\centering
\includegraphics[width=1.0\textwidth]{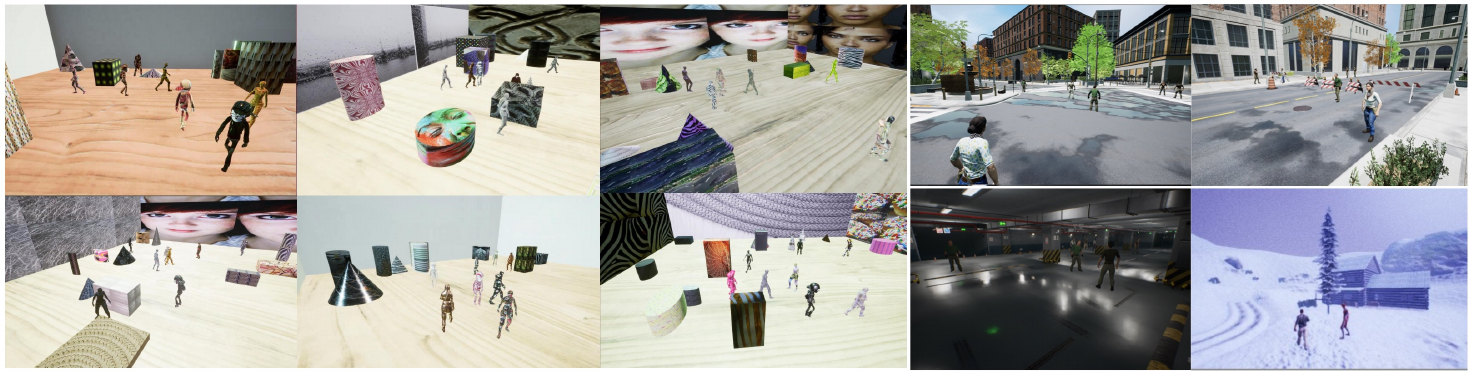}
\caption{Examples of the high-fidelity scenes for training and testing(from left to right): \emph{ComplexRoom}(the first three columns),   \emph{UrbanCity}(top in fourth column), \emph{UrbanRoad}(top in fifth column),\emph{Parking Lot}(down in fourth column), and \emph{SnowForest}(down in fifth column )}
\label{figs:test-env}
\end{figure}

\begin{figure}[h]
\centering
\includegraphics[width=1.0\linewidth]{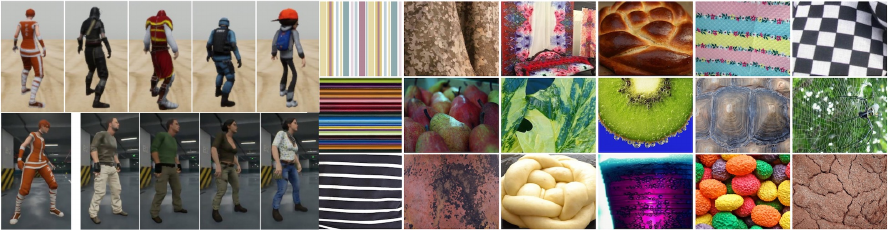}
\caption{\textbf{Left Top}: Five humanoid models used in \emph{Complex Room}. Their textures are randomized in training, using the pictures shown on the \textbf{Right Side}.
\textbf{Left Bottom}: The humanoid models used in Complex Room. The four humanoid models on the right are used in \emph{Parking Lot}, \emph{UrbanCity}, \emph{UrbanRoad}, and \emph{SnowForest}. Note that the red one is only used in \emph{Parking Lot (4D)} to produce multiple distractors with the same appearance as the target.
}
\label{figs:train-humanoid}
\end{figure}

\begin{figure}[h]
    \centering
    \includegraphics[width=0.5\linewidth]{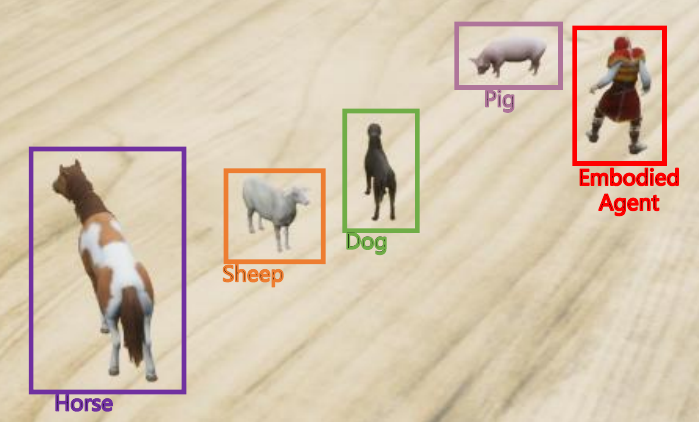}
    \caption{An Example of the multiple animal's environments for evaluating the generalization on the unseen category of target. The animals have different sizes and appearances, such as horses, dogs, sheep, and so on.}
    \label{fig:animal}
    \vspace{-0.5cm}
\end{figure}

We use high-fidelity environments from previous works\cite{zhong2023rspt,zhong2021distractor}, which are of diverse obstacles or distractors in the environment. A room-like environment (``Complex Room") is used for data collection or training baselines in an online manner. The other environments are used for evaluating the generalization and robustness of the agents. The advantages of training and evaluating in these environments are threefold. Firstly, all of them are constructed based on UnrealCV~\cite{qiu2017unrealcv} and simulate the physical properties of the real scene properly. Secondly, it provides information required for calculating long-term returns and evaluation metrics in experiments. Thirdly, it avoids the high cost of trial-and-error interaction in the real scene. We describe each environment in detail and demonstrate how they place challenges to active tracking tasks as below.

\textbf{\emph{Complex Room} and \emph{Simple Room}}. The Complex Room is a plain room used for training, with randomized obstacles, players, background textures, and illumination conditions. The left six images on Figure~\ref{figs:test-env} show the third-person view of the Complex Room. We use five humanoid models as targets and reset them every episode as shown in Figure~\ref{figs:train-humanoid}. Besides, to learn an appearance-invariant feature, we randomly choose pictures from a texture dataset \cite{Kylberg2011c} and set the textures of walls, floors, and players separately. Some examples are shown in Fig.~\ref{figs:train-humanoid}. To establish various illumination conditions, we randomize the intensity and color of each light source at the beginning of each episode. Additionally, in order to simulate the complex and challenging scenes, we shuffle twenty obstacles in \emph{Complex Room} at the beginning of each episode. Their shapes can be cubes, cones, spheres, and cylinders and their scales, textures, and locations are randomized.
The Simple Room is the clean version of the complex room for evaluating the most basic tracking ability of the agent. There are only four gray walls on the side, without any obstacles in the room. The other environmental factors are not randomized in the room.
To evaluate the generalization on the unseen target, we also introduce four animals (Horse, Sheep, Dog, and Pig) as the target and distractors in the Simple Room, as is shown in Figure~\ref{fig:animal}.

\textbf{\emph{Urban City.}} It is constructed to mimic a street intersection scene with tall buildings, street lights, stone piers, and some puddles which reflect other objects, as a consequence, it would be difficult to track a target there due to the unseen textures and shapes of objects. In this scene, we also test the distraction robustness by adding four distractors, which are of similar appearance to the targets. 

\textbf{\emph{Urban Road.}} Similar to UrbanCity, UrbanRoad is a street scene that is mainly composed of urban roads and roadblocks. What makes it more difficult for embodied visual tracking is that there are many different kinds of obstacles, most of which are unseen in the training environments.

\textbf{\emph{Parking Lot.}} This environment looks like an underground parking lot with uneven lighting conditions. While tracking in \emph{Parking Lot}, the tracker is required to recognize the target from a background with challenging illustrations and even track the targets occluded by some pillars. There are four unseen humanoids are used, illustrated in the ~Fig.~\ref{figs:train-humanoid}. This scene is also used for evaluating the distraction robustness. Note that the distractors and the target share the same humanoid model to make the distractions more challenging.

\textbf{\emph{Snow Forest.}} This is a typical snowy scene, where both the target and the surrounding scenery, such as trees, shrubs, and houses, are often veiled in snow fog.  It's important to note that in such conditions, first-person observations may experience impaired visibility due to snowflakes and fog, affecting the line of sight. Moreover, the undulating terrain adds to the difficulty of tracking.

Figure~\ref{figs:SimVis} shows four typical cases of our agents that successfully track the target in the virtual environment. We can see that the segmentation mask extracted by VFM (DEVA) is noisy in such complex cases, but the learned recurrent policy still can keep tracking. 
More vivid examples of the collected data and the tracking videos are shown in the demo page: \url{https://sites.google.com/view/offline-evt}.

\section{Offline Data Collection}
\label{sec:Noise Level}
In this section, we introduce the implementation details of the data collection process, including the state-based controller for the expert tracker, the noise action applied in data collection, and the distribution of the collected data.

\subsection{State-based Controller}
We adopt PID controllers to move an agent to follow a target object and keep the target in a specific distance and relative angle (in the front 3 meters) as follows:
\begin{itemize}
    \item Define the expected distance and angle as the setpoints for the PID controller. For example, the expected distance can be 3 meters and the desired angle can be 0 degrees (directly in front of the agent).
    \item Measure the actual distance and angle between the agent and the target object using the grounded state, which is accessible by the UnrealCV API. These are the process variables for the PID controller.
    \item Calculate the errors between the setpoints and the process variables. These are the inputs for the PID controller.
    \item Apply the PID equation to generate the control output, which is the speed and direction of the agent. The PID equation can be written as:
    \begin{equation}
        u(t)=K_p e(t)+K_i \int_0^t e(\tau) d \tau+K_d \frac{d e(t)}{d t}
    \end{equation}

where $u(t)$ is the control output, $e(t)$ is the error, $K_p$, $K_i$, and $K_d$ are the proportional, integral, and derivative gains, respectively.
    \item Adjust the gains to achieve the desired performance of the PID controller. For example, increasing $K_p$ will make the agent respond faster to the error, but may also cause overshoot or oscillation. Increasing $K_i$ will reduce the steady-state error, but may also cause integral windup or slow response. Increasing $K_d$ will reduce the overshoot and oscillation, but may also amplify the noise or cause a derivative kick.  Note that when the output is out of the range defined in the action space, we will clip it. The tuned gains are shown in Table~\ref{tab:pid}.
    
    \begin{table}[h]
        \centering
        \caption{The gains we used in the PID controller.}
        \begin{tabular}{cccc}
        \hline
        Controller & $K_p$ & $K_i$ & $K_d$ \\
        \hline
        Speed & 5 & 0.1 & 0.05\\
        Angle    & 1 & 0.01 & 0\\
        \hline
        \end{tabular}
        \label{tab:pid}
    \end{table}
\end{itemize}
\subsection{The Noise Level and Dataset Size}
\label{sup_data_noise}
In this section, we first present our detailed data size and noise level ablation result mentioned in Sec 5.6, shown in Table \ref{tab:noise}. Then, we introduce our strategies for incorporating noise actions. We set a threshold $p$, and if the probability value at time $t$ is greater than $p$, which is $P(t)>p$, the agent takes a random action from the action space and continues for $L$ steps. We also adopt a random strategy for setting the step length $L$, with an upper limit of 5. After random actions of the agent ends, we use a random function to determine the duration of the next random action $L$. Here, we set the upper limit to 5 because we found that the number of times the agent failed in a round significantly increased beyond 5. Therefore, we empirically set the upper limit of the random step length to 5.
\begin{table}[t]
    \centering
    \caption{Evaluating the effect of the data size and noise level in data collection in \textit{Complex Room}.}
    \begin{tabular}{cccr}
    \hline
       & Data Size & Noise Level &  Results (AR/EL/SR) \\ \hline
       Ours & 40k & 1 & 17/297/0.18 \\ 
       Ours & 40k & 2 & 282/473/0.82 \\ 
       Ours & 40k & 3 & 221/439/0.68 \\ 
       Ours & 40k & 4 & 123/373/0.45 \\ \hline
       Ours & 20k & 2 & 49/328/0.23 \\ 
       Ours & 80k & 2 & 263/454/0.76 \\ \hline
       BC & 200k  & 0 & 143/341/0.32 \\
       BC & 500k  & 0 & 199/371/0.45 \\
       BC & 1000k & 0 &  222/387/0.49\\ \hline
    \end{tabular}
        \vspace{-0.3cm}
    \label{tab:noise}
\end{table}
\begin{figure*}[t]
    \centering
    \includegraphics[width=\linewidth]{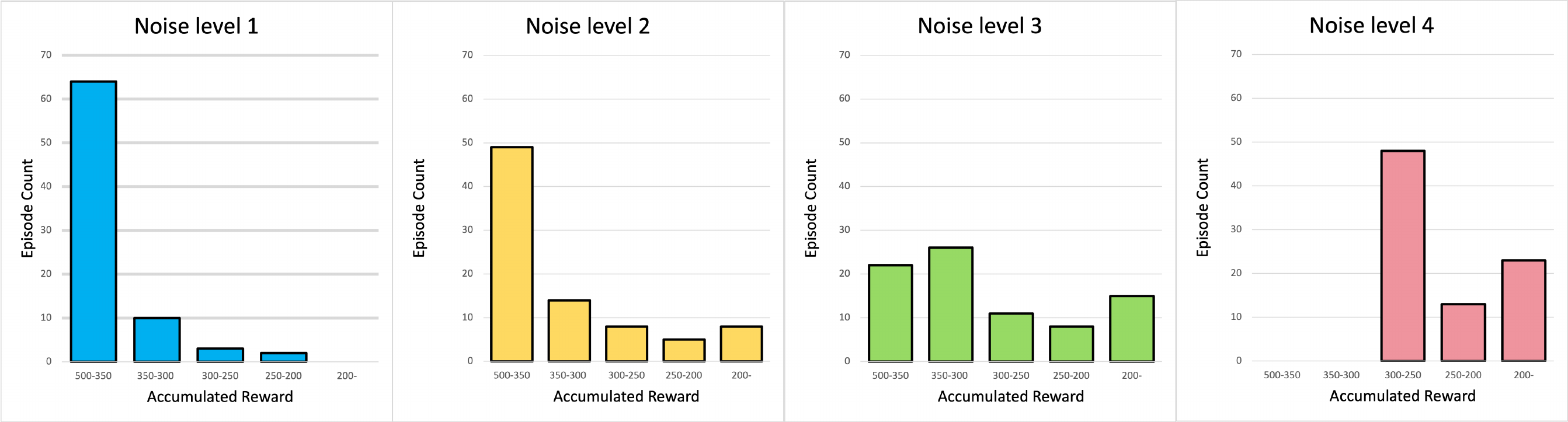}
    \caption{
    Comparison of data distribution for each noise level.  }
    \label{fig:Noise Data}
\end{figure*}

\subsection{Data Distribution}

We visualized the distributions of four different noise levels used in the experimental section. We defined the noise levels based on the distribution of cumulative rewards for each episode in each dataset. The range of cumulative rewards was divided into five categories: 500-350, 350-300, 300-250, 250-200, and below 200. The higher the cumulative reward, the fewer noise actions in this episode. We calculated the number of episodes belonging to different noise categories in four datasets, and their detailed distributions are shown in Figure~\ref{fig:Noise Data}. We can observe that as the noise level increases, the number of episodes containing high noise actions also increases. At noise level 4, all episodes are distributed in the range of cumulative rewards below 300.

\section{Implementation Details}
\subsection{Baselines}
We further explain the details of the baselines, including DiMP, BC methods, and other end-to-end methods. 

\begin{figure}[t]
    \begin{center}
    \includegraphics[width=0.6\linewidth]{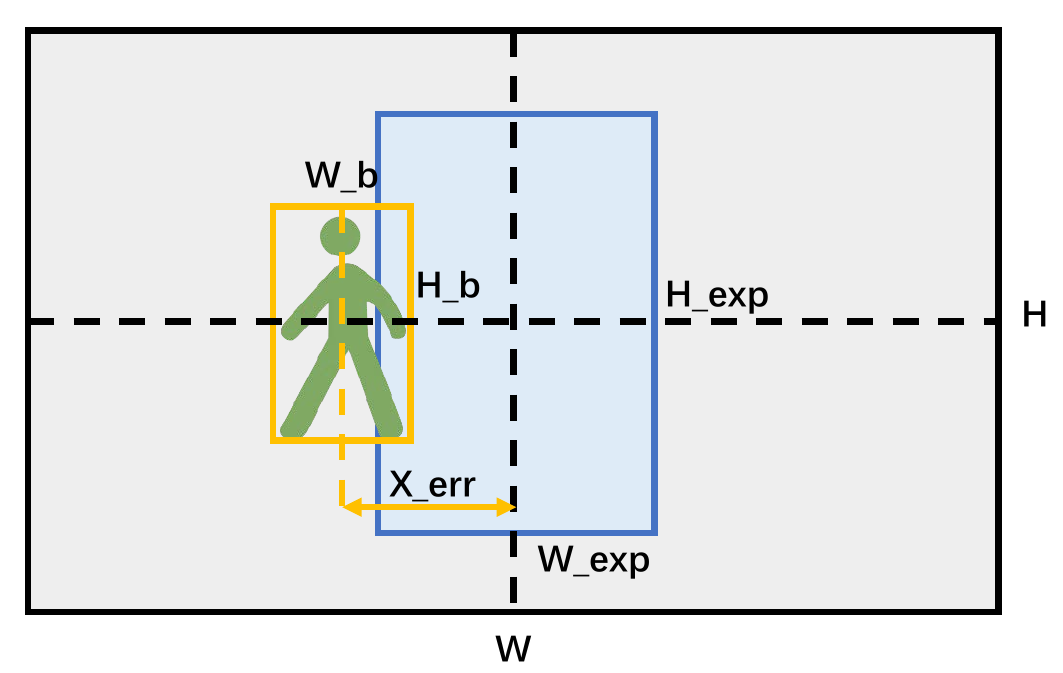}
    \end{center}
    \vspace{-0.4cm}
    \caption{An example to illustrate errors for the PID controller~\cite{zhong2021distractor}. $X_{err}$ measures how far the bounding box is horizontally away from the desired position. $X_{err}$ is negative, meaning that the camera should move left so the object is closer to the center of the image. The distance to the target is reflected by the size of the bounding box $S_{err}$. }
    \vspace{-0.3cm}
    \label{fig:pid}
\end{figure}

\emph{DiMP (Discriminative Model Prediction)} serves as an off-the-shelf two-stage tracker. It consists of a visual tracker and a PID controller. We utilize the official implementation provided by its repository at \url{https://github.com/visionml/pytracking}. The camera controller outputs actions based on the specific error of the bounding box, provided by the visual tracker. The controller chooses an action from the action space according to the horizontal error $X_{err}$ and the size error $S_{err} = W_b\times H_b - W_{exp}\times H_{exp}$, shown as Figure~\ref{fig:pid}. 

\emph{Behavior Cloning (BC)} is an offline reinforcement learning (RL) method akin to supervised learning, utilizing labeled data (expert data) for training. In our implementation, for a fair comparison, we employ the same visual representation (text-condition mask generated by VFM and re-targeting mechanism) as input for the BC methods. However, we diverge by employing pure data as training data, aiming for the tracker to learn the tracking policy under expert policy supervision. While the offline training strategy ensures high training efficiency akin to our proposed method, it is susceptible to overfitting due to the limited coverage of conditions in the dataset, potentially resulting in weak generalization.

\emph{End-to-end Methods}, including SARL, AD-VAT, AD-VAT+, and TS, follow the networks used in ~\cite{zhong2021distractor}. We implement them based on the official implementation in \url{https://github.com/zfw1226/active_tracking_rl/distractor}. We strictly follow the hyper-parameters used in the original version. $4$ parallel workers are launched during training. 

\subsection{VFMs}
In this paper, we employ 4 different VFMs (DEVA, SAM-Track, DINOv2 and E-SAM) to generate visual representation.

\emph{DEVA} extends the Segment Anything Model (SAM) to video, integrating Grounding-DINO with SAM for text-prompted open-world video segmentation of novel objects. It provides pixel-wise segmentations and realizes zero-shot transferring to open-vocabulary objects. We utilize the official implementation provided by \url{https://github.com/hkchengrex/Tracking-Anything-with-DEVA}. 

\emph{SAM-Track} combines the interactive key-frame segmentation model SAM with the proposed AOT-based tracking model (DeAOT), which can dynamically detect and segment new objects. We utilize the official implementation provided by \url{https://github.com/z-x-yang/Segment-and-Track-Anything}.

\emph{DINOv2} is trained via a novel approach for computer vision tasks, harnesses the power of self-supervised learning. Unlike traditional methods that rely on labeled data, DINOv2 trains directly from images, bypassing the need for text-image pairs. By doing so, it acquires rich information about input images, making it versatile for a wide range of tasks. DINOv2 features a Vision Transformer (ViT) head that readily computes embeddings for image-to-image retrieval, classification tasks, segmentation, and depth estimation. 
In this paper, we utilized its segmentation function. We use the official implementation \url{https://github.com/facebookresearch/dinov2}.

\emph{Efficient-SAM} adopts the self-supervised MAE method. It masks out part of the image, then restores it, and compares it with the features generated by the original SAM's ViT-h for contrastive learning. Ultimately, it also replaces the ViT module with a lightweight encoder module,  successfully improving inference speed. We use the official implementation \url{https://github.com/xetdata/EfficientSAM}.

\subsection{Comparing the Computation Cost of Different VFMs}
\label{inference time}

 \begin{table}[h]
\centering
\caption{The inference time of deploying various VFMS in our model. All configurations are run on Nvidia GTX 3090.}
\begin{tabular}{l|ccc}
\hline\hline
  Name     & Average Inference Time (ms/step) & GPU memory & AR/EL/SR   \\ \hline
  DEVA   & $66 $ & 8280MB &306/479/0.92    \\ 
  SAM-Track & $100$ & 1730MB& 339/497/0.97\\
  DINOv2 &  $100$ & 1130MB&234/434/0.74\\
  E-SAM & $256$ &1800MB& -34/253/0.19\\ \hline
\end{tabular}

\label{time}
\end{table}

\textbf{Real-time inference} is crucial for embodied visual tracking. In Table~\ref{time}, we present the average inference times for different VFMs. DEVA, SAM-Track, and DINOv2 are well-suited for real-time inference, with DEVA and SAM-Track demonstrating superior efficiency. 

\subsection{Offline RL}
Our CQL implementation is based on a Github Repository: ~\url{https://github.com/BY571/CQL}. We modify the data sampler and the optimization process to support the RNN-based recurrent policy. The hyper-parameters that we used in offline reinforcement learning and the neural network structures are listed in Table~\ref{HyperParam} and Table~\ref{cnn}.


\begin{table}[ht]
\centering
\caption{The hyper-parameters used for offline training and the policy network.}
\begin{tabular}{l|cc}
\hline\hline
Name                 & Symbol           & Value \\ \hline
Learning Rate       & $\alpha$         &  3e-5\\ \hline
Discount Factor     & $\gamma$         &  0.99    \\ \hline
Batch Size           & -         &  32\\ \hline
LSTM update step      & -         & 20 \\ \hline
LSTM Input  Dimension & - & 256 \\ \hline
LSTM Output Dimension & - & 64 \\ \hline
LSTM Hidden Layer size & - & 1 \\ \hline
\end{tabular}

\label{HyperParam}
\end{table}

\begin{table}[ht]
\centering
\caption{The neural network structure, where 8$\times$8-16S4 means 16 filters of size 8$\times$8 and stride 4, FC256 indicates fully connected layer with dimension 256, and LSTM64 indicates that all the sizes in the LSTM unit are 64.}
\begin{tabular}{c|c|c|c|c|c|c}
\hline\hline
Module & \multicolumn{3}{|c|}{Mask Encoder} & Temporal Encoder & Actor & Critic \\ \hline
Layer\# & CNN & CNN & FC & LSTM & FC & FC\\ \hline
Parameters &     8$\times$8-16\emph{S}4     &       4$\times$4-32\emph{S}2       &      256     &       64       &  2 &2 \\
 \hline
\end{tabular}
\label{cnn}

\end{table}